\relax
\documentclass[letterpaper]{article} 
\usepackage{aaai20}  
\usepackage{times}  
\usepackage{helvet} 
\usepackage{courier}  
\usepackage[hyphens]{url}  
\usepackage{graphicx} 

\usepackage{xspace}
\newtheorem{defn}{Definition}
\usepackage{times}
\usepackage{latexsym}
\usepackage{graphicx} 
\usepackage{url}
\usepackage{multirow}
\usepackage{amsfonts}
\usepackage{amssymb}
\usepackage{color}
\usepackage{amsmath}
\usepackage{algorithm}
\usepackage{algorithmicx}
\usepackage{algpseudocode}
\usepackage{mathtools}
\newcommand{\newcite}[1]{\citeauthor{#1} \shortcite{#1}}

\newcommand{\modelname}{Span Keyphrase Extraction~\xspace}

\usepackage{graphicx}  
\title{Keyphrase Extraction with Span-based Feature Representations}
\author{
\normalsize{Funan Mu$^\sharp$, Zhenting Yu$^\sharp$, LiFeng Wang$^\sharp$, Yequan Wang$^\sharp$, Qingyu Yin$^\flat$, Yibo Sun$^\flat$, Liqun Liu$^\sharp$, Teng Ma$^\sharp$, Jing Tang$^\sharp$, Xing Zhou$^\sharp$} \\
$^\sharp$Tencent\\
$^\flat$Harbin Institute of Technology, China\\
{\tt \{marvinmu, trumpyu, fandywang, yequanwang, liqunliu, neoma, jamesjtang, leostarzhou\}@tencent.com} \\
  {\tt \{yqy, ybsun\}@ir.hit.edu.cn} \\
}

\begin{document}
\maketitle

\begin{abstract}
Keyphrases are capable of providing semantic metadata characterizing documents and producing an overview of the content of a document. Since keyphrase extraction is able to facilitate the management, categorization, and retrieval of information, it has received much attention in recent years.
There are three approaches to address keyphrase extraction: (i) traditional two-step ranking method, (ii) sequence labeling and (iii) generation using neural networks. 
Two-step ranking approach is based on feature engineering, which is labor intensive and domain dependent. Sequence labeling is not able to tackle overlapping phrases. Generation methods (i.e., Sequence-to-sequence neural network models) overcome those shortcomings, so they have been widely studied and gain state-of-the-art performance. However, generation methods can not utilize context information effectively. 
In this paper, we propose a novelty \modelname model that extracts span-based feature representation of keyphrase directly from all the content tokens. In this way, our model obtains representation for each keyphrase and further learns to capture the interaction between keyphrases in one document to get better ranking results. In addition, with the help of tokens, our model is able to extract overlapped keyphrases. Experimental results on the benchmark datasets show that our proposed model outperforms the existing methods by a large margin.
\end{abstract}

\section{Introduction}

Keyphrases are the most important critical and topical phrases for a given text. They are capable of providing a concise summary for a piece of text, and facilitating the management, categorization, and retrieval of information. Keyphrase extraction~\cite{witten2005kea} is widely used in real-world applications such as recommendation system \cite{ferrara2011keyphrase} and information retrieval~\cite{DBLP:conf/amcis/LiWBC04}.
    
Several methods have been proposed in previous studies. There are three main approaches: (i) traditional two-step ranking method, (ii) sequence labeling method and (iii) generation-based neural network method. A straightforward way of keyphrase extraction is to decompose this task into two steps: candidate phrases generation and candidate phrases scoring \cite{witten2005kea,medelyan2009human}.
In the first step, models generate a list of candidate phrases using the N-grams or phrases with certain part-of-speech patterns. In the second step, candidate phrases are scored by its probability of being a keyphrase in the given document. These two-stage ranking-based methods treat candidate phrases individually, which makes it almost impossible to capture the contextual information of candidates and interactions between different phrases. Further, existing two-stage methods \cite{witten2005kea,medelyan2009human} are based on feature engineering, which is labor intensive and domain dependent. 
Another intuitive approach is to regard keyphrase extraction as a sequence labeling task~\cite{zhang2016keyphrase}. However, sequence labeling approach can hardly tackle keyphrase with overlapping words. 
\begin{figure}
    \centering
    \includegraphics[width=6.5cm]{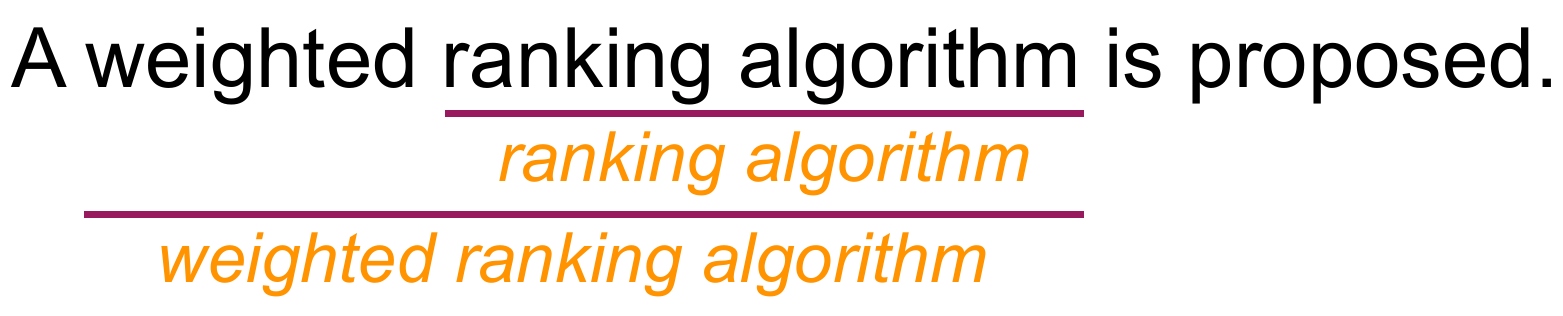}
    \caption{The overlap phenomenon in keyphrases.}
    \label{fig:overlap}
    \vspace{-3ex}
\end{figure}
As shown in Figure~\ref{fig:overlap}, ``weighted ranking algorithm'' and ``ranking algorithm'' are both keyphrases that provide semantic information of different granularity. Unfortunately, sequence labeling methods are not able to extract both of them at the same time.
With the development of deep learning, especially sequence-to-sequence methods, generation-based methods~\cite{meng2017deep,chen2018keyphrase} have attracted much attention. Admittedly, generation-based approach is capable of dealing with the overlap keyphrases without much labor-intensive feature engineering, but such an approach has two shortcomings. Firstly, generation method can not utilize context information effectively and capture the interaction of phrases. Secondly, generation method produces tokens of keyphrases continuously which suffers the problem of not utilizing phrases-level information. In other words, tokens in a phrase can only make sense when they appear as a whole sequence. Therefore, a desirable solution should be able to capture the information within word sequences and take advantage of this span-based information when predicting the keyphrases.

In this paper, we make the very first step to perform keyphrase extraction by~\modelname ~(SKE) model based on span-based feature representation. The proposed SKE model first extracts candidate phrases using the certain part-of-speech patterns~\cite{le2016unsupervised} and records the beginning and ending positions of each candidate phrase as \textbf{spans}. After that, Bert~\cite{devlin2018bert} or recurrent neural network based on word vectors is used to represent the high-level concept of phrases. We call the high-level representation span-based representation. Afterward, bidirectional recurrent neural network (i.e., LSTM and GRU) is used to capture the interaction of span-based representation to get higher-level phrase representation. After getting the phrase representation, we are able to use them to classify the candidate phrase. 
Especially, BERT is pre-trained with a large amount of data and contains language knowledge, position information, and contextual information, so it is designed to encode the phrase representation. The generation of candidate phrases allows the overlap between phrases. Further, the ranking processing is able to utilize the context information of phrases. 
In summary, with the well-designed model, \modelname can obtain the overall semantic meaning of both the documents and keyphrases and learn the interaction between phrases.
The main contributions are as follows:

\begin{itemize}
    \item To the best of our knowledge, \modelname is the first attempt to use span-based features for keyphrase extraction. The span-based approach is capable of tackling overlap problem effectively. 
    \item The proposed \modelname model is capable of utilizing context information. Span-based features of phrases are based on the document. Bidirectional recurrent network (i.e., LSTM and GRU), which can utilize context information, are employed to capture the interactions between phrases based on span-based representations. 
    \item We conduct experiments on five benchmark datasets, comparing our SKE model with strong baseline models. Experiment results show that our proposed model achieves state-of-the-art performance.
\end{itemize} 

\section{Related Work} \label{relatedwork}
    Obtaining high-quality keyphrases for documents is a classic and challenging problem in natural language processing, which has been widely studied in previous works. The existing methods can be categorized into three groups: two-step ranking approach, sequence labeling approach and generation approach.

    Traditional ranking-based method consists of two stages. The first stage is to acquire a set of candidate phrases with heuristic methods, such as extracting important n-grams \cite{hulth2003improved} and selecting text chunks with certain part-of-speech tags \cite{liu2011automatic,le2016unsupervised}. These candidates ought to cover the correct answer as much as possible since the coverage will heavily affect the final result. The second stage is to score the candidate phrases through the probability of being a keyphrase using classifier or heuristic metrics. Keyphrases are selected from the top-ranked candidates. \newcite{witten2005kea,hulth2003improved,medelyan2009human,zhang2019keywords} use a classification model to solve this problem by features and machine learning models. These models are based on feature engineering, which is labor intensive and domain dependent.
    Other researchers look forward to unsupervised methods \cite{mihalcea2004textrank,liu2010automatic,zhang2013wordtopic}. On the other side, \newcite{tomokiyo2003language} apply two statistical language models to measure the phraseness and informativeness for phrases. \newcite{liu2011automatic} use a word alignment model, which learns a translation from the documents to the keyphrases. This approach alleviates the problem of vocabulary gaps between source and target to a certain degree whereas failed to handle semantic meaning.

    Another straightforward way is to regard keyphrase extraction as a sequence labeling task, \newcite{zhang2016keyphrase} propose a joint-layer recurrent neural network model to extract keyphrases from tweets. However, this method fails to handle overlapped keyphrases.

    With the successful application of the sequence-to-sequence model in the field of machine translation, keyphrase generation~\cite{meng2017deep,chen2018keyphrase} methods have received much attention. By attention mechanism \cite{bahdanau2014neural}, copy mechanism \cite{gu2016incorporating}, coverage mechanism \cite{tu2016modeling} and review mechanism \cite{chen2018keyphrase}, generation based models achieve the state-of-the-art performance.
    However, generation method can not utilize context information effectively and the phrases-level information.

    Our proposed \modelname is basically a two-stage ranking-based approach. The main contribution of our model is that the span-based approach is capable of tackling overlap problem effectively and is capable of utilizing context information. Previous two-stage methods \cite{witten2005kea,medelyan2009human} view each phrase as an instance even they belong to one document. However, we regard each document as an instance and use Bi-LSTM to capture the interaction between keyphrases to get better ranking results. Further, we use BERT \cite{devlin2018bert}, which is a pre-trained language representation, to take advantage of the abundant language knowledge, position information, and contextual information. 

\section{Methodology}\label{Methodology}
    This section gives the detail of our proposed \modelname model. First, we define the task of keyphrase extraction. Then, we give a brief presentation of the token features. Finally, we describe the method to extract span-based feature representations of phrases and how to rank them.

\normalsize

\begin{figure}[t]
    \centering
    \includegraphics[scale=0.71]{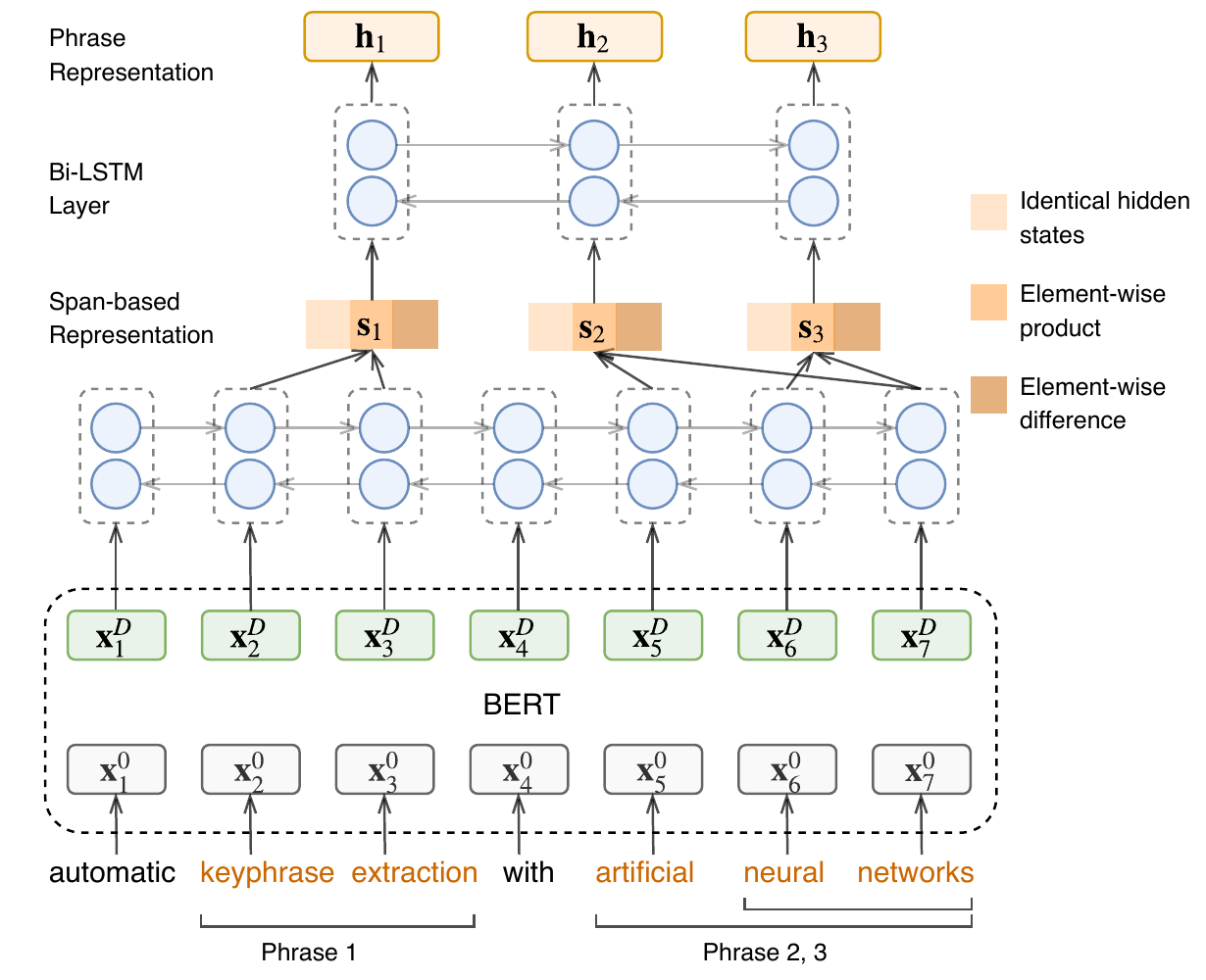}
    \caption{Model Architecture. We first use BERT to extract token features. Then span-based feature representations of phrases are produced based on token features. Another Bi-LSTM is adopted to learn the interactions of phrases.}
    \label{model_architecture}
\end{figure}

\subsection{Preliminaries}
    Given a keyphrase extraction dataset $\mathcal{D} = {\{\textbf{d}^{i},\textbf{p}^{i}\}}^{N}_{i=1}$, where $\textbf{d}^{i}$ is the $\mathnormal{i}$-th document text and $\textbf{p}^{i} = {\{\textbf{p}^{i,j}\}}^{N_{i}}_{j=1}$ is the keyphrase set of $\textbf{d}^{i}$. $N$ and $N_{i}$ are the number of documents and keyphrases of $\textbf{d}^{i}$ respectively. Both the document text and target keyphrase are token sequences which can be denoted as $\textbf{d}^{i} = \{t^{i}_{j}\}^{L^i}_{j=1}$ and $\textbf{p}^{i,j} = \{t^{i,j}_{j}\}^{K^{i,j}}_{j=1}$ . $L^{i}$ and $K^{i,j}$ denote the length of token sequences of $\textbf{d}^{i}$ and $\textbf{p}^{i,j}$ respectively. 
  
    We denote the beginning and ending position of phrases in the document as spans and define the span-based feature as follows: 
 \begin{defn}{\textbf{[Span-based Feature]}}
    Let $\textbf{T}=\textbf{t}_{1},\textbf{t}_{2},\dots,\textbf{t}_{L}$ be the features of tokens in a document. The span of a phrase is $\{b, e\}, 1 \leq b, e \leq L$. The span-based feature are the relations between $\textbf{t}_b$ and $\textbf{t}_e$.
  \label{define:SF}
\end{defn}
  
    According to the definition, we apply three relations, i.e., identical, element multiply and element difference. Details are described in \textbf{Span-based Feature Representations} Section.

    We extract candidate phrases set of $\textbf{d}^{i}$ as $\textbf{c}^{i}$. Number of $\textbf{c}^{i}$ is denoted as $M^{i}$. Intersection phrases of $\textbf{c}^{i}$ and $\textbf{k}^{i}$ are the positives denoted as $\textbf{c}^{i+}$. The rest candidate phrases are the negatives denoted as $\textbf{c}^{i-}$. In this way we can convert keyphrase extraction to a binary classification task by minimizing the cross entropy loss
\begin{equation}
    -\sum_{i=1}^{N}[\sum_{p\in\textbf{c}^{i+}}\log h(p)+\sum_{n \in\textbf{c}^{i-}}\log (1-h(n))], \label{classification_equation}
\end{equation}
   where $h(x)$ is the likelihood of being a keyphrase. 
   
   Hinge loss is another effective objective  to optimize our proposed model:
\begin{equation}
    \sum_{i=1}^{N}\sum_{p\in\textbf{c}^{i+}}\sum_{n\in\textbf{c}^{i-}} \max(0, m - (h(p) - h(n))), \label{ranking_equation}
\end{equation}
where $m$ is the shorthand of margin which is a hyperparameter.

Figure \ref{model_architecture} gives the detail of \modelname model. The bottom of the figure is BERT, which is used to extract token features. The middle part of the figure is the spaned-based representation module. The top of the figure is the bidirectional recurrent neural network, which is used to learn the interaction of phrases.  More details about the processing  is given in Algorithm \ref{alg:A}.

\begin{algorithm}[t]
    \caption{Training procedure of the proposed model}
    \label{alg:A}
    \footnotesize
    \begin{algorithmic}[1]
    \Require The train dataset $\mathcal{D}$; The BERT model $\varphi_{B}$; The part-of-speech tagger $pos_t$; The part-of-speech pattern $pos_p$; The Porter Stemmmer $stemmer$;
    \item \textbf{for} each (\textbf{d}, $\textbf{p}=(\textbf{p}^{1},...,\textbf{p}^{k}))$ $\in\mathcal{D}$ \textbf{do}
    \item $\quad$ tagger document $\textbf{d}_{pos} = pos_{t}(\textbf{d})$;
    \item $\quad$ extract candidate $\textbf{c} = pos_p(\textbf{d}_{pos})$ 
    \item $\quad$ record spans of $\textbf{c}$ as $span(\textbf{c})$;
    \item $\quad$ stem candidate phrases $c_{s} = stemmer(\textbf{c})$;
    \item $\quad$ stem keyphrases $p_{s} = stemmer(\textbf{p})$;
    \item $\quad$ $\textbf{c}^{+}=\textbf{c}_{s}\cap\textbf{p}_{s}$;
    \item $\quad$ $\textbf{c}^{-}=\textbf{c}_{s}-\textbf{c}_{s}^{+}$;
    \item $\quad$ compute token features $X^{D}=\varphi_{B}(\textbf{d})$;
    \item $\quad$ $\overrightarrow{\textbf{t}_{i}} = \overrightarrow{LSTM}(\textbf{x}^{D}_{i}), i \in [1, L]$;
    \item $\quad$ $\overleftarrow{\textbf{t}_{i}} = \overleftarrow{LSTM}(\textbf{x}^{D}_{i}), i \in [L, 1]$;
    \item $\quad$ \textbf{for} each (b, e) $\in span(\textbf{c})$; \textbf{do}
    \item $\quad\quad$ $\textbf{p}=(\overrightarrow{\textbf{t}_{b}}, \overrightarrow{\textbf{t}_{e}},\overleftarrow{\textbf{t}_{b}},\overleftarrow{\textbf{t}_{e}},\overrightarrow{\textbf{t}_{b}} \ast \overrightarrow{\textbf{t}_{e}}, \overleftarrow{\textbf{t}_{b}} \ast \overleftarrow{\textbf{t}_{e}},\overrightarrow{\textbf{t}_{e}} - \overrightarrow{\textbf{t}_{b}}, \overleftarrow{\textbf{t}_{b}} - \overleftarrow{\textbf{t}_{e}}
    )$;
    \item $\quad\quad$ $\textbf{h}_{i} = (\overrightarrow{LSTM}(\textbf{p}_{i}), \overleftarrow{LSTM}(\textbf{p}_{i})), i \in [1, M]$;
    \item $\quad$ \textbf{end for}
    \item $\quad$ compute $loss$ for $\textbf{c}^{+}$ and $\textbf{c}^{-}$ with $\textbf{H}$;
    \item $\quad$ compute $gradient$ and $update$;
    \item \textbf{end for}
    \end{algorithmic}
\end{algorithm}

\subsection{Token Features}\label{bert_section}
    As mentioned before, we first obtain the features of tokens with BERT, which utilizes the abundant language knowledge, position information, and contextual information it contains. Given $\textbf{d} = \{t_{i}\}^{L}_{i}$, BERT begins by converting the sequence of tokens into a sequence of vectors $\textbf{X}^{0} = \{\textbf{x}^{0}_{i}\}^{L}_{i}$, $\textbf{x}^{0}_{i} \in \mathbb{R}^{d}$. Each of these vectors is the sum of a token embedding, a positional embedding that represents the position of the token in the sequence, and a segment embedding that represents whether the token is in the source text or the auxiliary text. We only have source text so the segment embeddings are the same for all tokens. Then several Transformer \cite{vaswani2017attention} layers are applied to get the final representations. Each Transformer layer has two sub-layers. The first sub-layer is a multi-head self-attention mechanism \cite{lin2017structured,yin2018zero}, and the second sub-layer is a simple, position-wise fully connected feed-forward network. A residual connection \cite{he2016deep} is employed around each of the two sub-layers, followed by layer normalization \cite{lei2016layer}. 
    By concatenating $h$ head self-attention output, we obtain the final values of multi-head self-attention:
\begin{equation}
\begin{aligned}
    mh(X^{i}) &=Concat(head_{1}, ..., head_{h})W^{O} \\
    H_{i} &= {\rm LayerNorm}(mh(X^{i}) + X^{i}) 
\end{aligned}
\end{equation}
\begin{displaymath}
\begin{aligned}
{\rm where\ } &head_{j} = Att(X^{i}W^{Q}_{j}, X^{i}W^{K}_{j}, X^{i}W^{V}_{j}) \\
&Att(Q, K, V) = {\rm softmax}(\frac{QK^{T}}{\sqrt{d}})V
\end{aligned}
\end{displaymath}

    This is followed by a fully connected feed-forward network with GELU \cite{hendrycks2016gaussian} activation
\begin{equation}
    X^{i+1} = {\rm LayerNorm(GELU}(H_{i}U)V + H_i)
\end{equation}
    We use the final hidden output of BERT $X^{D} \in \mathbb{R}^{L \times d}$ as the representations of corresponding tokens. 

\begin{figure}[t]
    \vspace{-2em}
    \centering
    \includegraphics[scale=0.6]{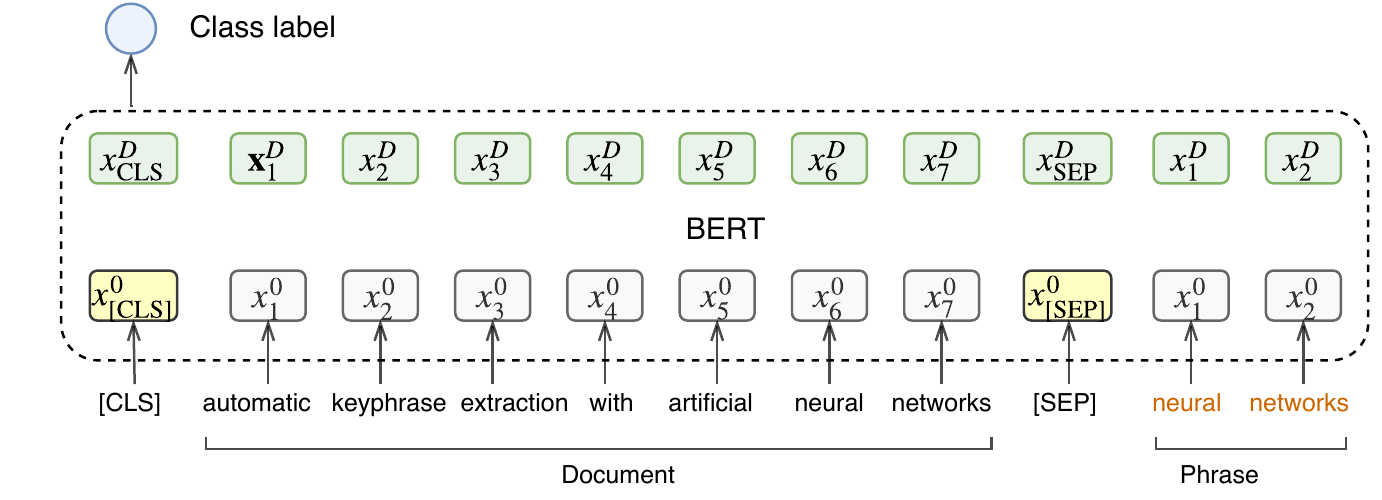}
    \vspace{-1em}
    \caption{Sentence pair classfication base on BERT. We use the final hidden vector $\textbf{x}^{D}_{[CLS]}$ as the aggregate representation. }
    \label{bert_pair}
\end{figure}

\subsection{Span-based Feature Representations}\label{span_features}
    We extract the spans of phrases and use $X^{D}$ produced by BERT  to extract features of phrases. Firstly, a Bi-LSTM model takes the token's hidden representation $X^{D}$ as input to get local-aware features of tokens. The hidden dimension of the Bi-LSTM model is $d/2$.
\begin{equation}
\begin{aligned}
	\overrightarrow{\textbf{t}_{i}} = \overrightarrow{LSTM}(\textbf{x}^{D}_{i}), i \in [1, L] \\
	\overleftarrow{\textbf{t}_{i}} = \overleftarrow{LSTM}(\textbf{x}^{D}_{i}), i \in [L, 1]
\end{aligned}
\end{equation}
    We denote the forward representations of the beginning and the ending tokens as $\overrightarrow{\textbf{t}_{b}}$ and $\overrightarrow{\textbf{t}_{e}}$, the backward representations of the beginning and the ending tokens as  $\overleftarrow{\textbf{t}_{b}}$ and $\overleftarrow{\textbf{t}_{e}}$. Inspired by \newcite{conneau2017supervised}, we concatenate three kinds of vectors as the phrase representation: (i) identical representations $(\overrightarrow{\textbf{t}_{b}}, \overrightarrow{\textbf{t}_{e}}, \overleftarrow{\textbf{t}_{b}}$, $\overleftarrow{\textbf{t}_{e}})$; (ii) element-wise product ($\overrightarrow{\textbf{t}_{b}} \ast \overrightarrow{\textbf{t}_{e}}, \overleftarrow{\textbf{t}_{b}} \ast \overleftarrow{\textbf{t}_{e}})$; and (iii) element-wise difference ($\overrightarrow{\textbf{t}_{e}} - \overrightarrow{\textbf{t}_{b}}, \overleftarrow{\textbf{t}_{b}} - \overleftarrow{\textbf{t}_{e}}$). In this way, we get span-based feature representations $\textbf{S} \in \mathbb{R}^{M \times 4d}$ of phrases. Then, the representations of phrases are feed into another Bi-LSTM with a $2d$ hidden dimension to learn the interaction between the phrases to get better ranking results. By concatenating the forward and backward vectors, we get the final representation $\textbf{H} \in \mathbb{R}^{M \times 4d}$.
\begin{equation}
\begin{aligned}
    \overrightarrow{\textbf{h}_{i}} &= \overrightarrow{LSTM}(\textbf{s}_{i}), i \in [1, M] \\
    \overleftarrow{\textbf{h}_{i}} &= \overleftarrow{LSTM}(\textbf{s}_{i}), i \in [M, 1] \\
    \textbf{h}_{i} &= (\overleftarrow{\textbf{h}_{i}}, \overrightarrow{\textbf{h}_{i}}), i \in [1, M]
\end{aligned}
\end{equation}

\subsection{Ranking Phrases}\label{rank_phrase}
    We propose to use two methods to train the model.
    The first method is regarding the task as a binary classification problem. We use a fully connected feed-forward network to obtain the final representation of phrase denoted as $\textbf{L} \in \mathbb{R}^{M \times 2}$. We conduct softmax on $\textbf{L}$ and use the cross entropy loss shown in Equation (\ref{classification_equation}).
    The second way is regarding the task as a ranking problem. We use a fully connected feed-forward network with sigmoid activation to obtain the final representation of phrase $\textbf{L} \in \mathbb{R}^{M \times 1}$. Equation (\ref{ranking_equation}) gives the detail of  hinge loss.

\begin{table*}[t]
    \centering
    \normalsize
    \scalebox{0.9}{
    \begin{tabular}{c|c|c|c|c|c|c|c}
    \hline
    \hline
        \textbf{Dataset} & \textbf{Inspec} & \textbf{Krapivin} & \textbf{NUS} & \textbf{SemEval} & \textbf{KP20k} & \textbf{KP Training} & \textbf{KP Validation} \\
    \hline
    \hline
        \textbf{\#$\textbf{c}^{i+}$} & 6.16 & 2.80 & 4.87 & 5.26 & 2.86 & 2.86 & 2.85\\
    \hline
        \textbf{\#Keyphrase} & 7.13 & 3.25 & 6.31 & 6.25 & 3.16 & 3.14 & 3.15\\
    \hline
        \textbf{Keyphrases Coverage by $\textbf{c}^{i+}$} & 86.36\% & 86.14\% & 77.16\% & 84.16\% & 90.50\% & 90.58\% & 90.58\% \\
    \hline
        \textbf{\#$\textbf{c}^{i-}$} & 54.43 & 68.21 & 101.98 & 101.18 & 73.40 & 76.40 & 76.40 \\ 
    \hline
        \textbf{Ratio of \#$\textbf{c}^{i-}$ and \#$\textbf{c}^{i+}$} & 8.84 & 24.38 & 20.95 & 19.23 & 25.65 & 28.70 & 25.76 \\ 
    \hline
    \hline
    \end{tabular}}

    \caption{\label{font-table} 
    Details of candidate phrases and keyphrases. The coverage show that 9.42\% to 22.84\% keyphrases are not covered by candidate phrases. This will limit the final performance of our model. The ratio of $\textbf{c}^{i-}$ and $\textbf{c}^{i+}$ is range from 9.85 to 28.85 meaning that the classification task is imbalanced.}
    \label{number_of_cplus_and_keyphrases}
\end{table*}

\section{Experiments}\label{Experiment}
    This section begins by discussing the datasets we experiment on and the details of how we pre-process the datasets, followed by the baselines and the evaluation metrics. Then we present results and analysis.
\begin{table*}[t]
    \centering
    \normalsize
    \scalebox{0.9}{
    \begin{tabular}{c|c c|c c|c c|c c|c c}
    \hline
    \hline
        \multirow{2}{*}{\textbf{Method}} & \multicolumn{2}{|c|}{\textbf{Inspec}} & \multicolumn{2}{|c|}{\textbf{Krapivin}} & \multicolumn{2}{|c|}{\textbf{NUS}} & \multicolumn{2}{|c|}{\textbf{SemEval}} & \multicolumn{2}{|c}{\textbf{KP20k}} \\ 
        & $\textbf{F}_1\textbf{@5}$ & $\textbf{F}_1 \textbf{@10}$ 
        & $\textbf{F}_1\textbf{@5}$ & $\textbf{F}_1 \textbf{@10}$ 
        & $\textbf{F}_1\textbf{@5}$ & $\textbf{F}_1 \textbf{@10}$ 
        & $\textbf{F}_1\textbf{@5}$ & $\textbf{F}_1 \textbf{@10}$ 
        & $\textbf{F}_1\textbf{@5}$ & $\textbf{F}_1 \textbf{@10}$ \\
    \hline
    \hline
        \textbf{Tf-Idf} & 0.221 & 0.313 & 0.129 & 0.160 & 0.136 & 0.184 & 0.128 & 0.194 & 0.108 & 0.134 \\
    \hline
        \textbf{TextRank} & 0.223 & 0.281 & 0.189 & 0.162 & 0.195 & 0.196 & 0.176 & 0.187 & 0.180 & 0.150 \\
    \hline
    \hline
        \textbf{Maui} & 0.040 & 0.042 & 0.249 & 0.216 & 0.249 & 0.268 & 0.044 & 0.039  & 0.273 & 0.240 \\
    \hline
        \textbf{KEA} & 0.098 & 0.126 & 0.123 & 0.134 & 0.069 & 0.084 & 0.025 & 0.026 & 0.182 & 0.167 \\
    \hline
    \hline
        \textbf{CopyRNN} & 0.278 & 0.341 & 0.311 & 0.266 & 0.334 & 0.326 & 0.293 & 0.304 & 0.333 & 0.262 \\
    \hline
        \textbf{CorrRNN} & - & - & \textbf{0.318} & \textbf{0.278} & 0.361 & 0.335 & 0.320 & 0.320 & - & - \\
    \hline
    \hline
        \textbf{BERT-Base-Pair} & 0.302 & 0.340 & \text0.288 & 0.247 & 0.382 & 0.362 & 0.316 & 0.330 & 0.373 & 0.313 \\
    \hline
        \textbf{SKE-Base-Rank} & 0.289 & 0.321 & 0.287 & 0.236 & 0.389 & 0.365 & 0.354 & 0.337 & 0.381 & 0.324 \\
    \hline
        \textbf{SKE-Base-Cls} & \textbf{0.305} & \textbf{0.342} & 0.312 & 0.251 & 0.395 & 0.371 & 0.352 & 0.342 & 0.386 & 0.326 \\
    \hline
    \hline
        \textbf{SKE-Large-Rank} & 0.300 & 0.334 & 0.313 & 0.264 & 0.400 & \textbf{0.379} & 0.356 & 0.351 & \textbf{0.392} & 0.328 \\
    \hline
        \textbf{SKE-Large-Cls} & 0.294 & 0.334 & 0.309 & 0.252 & \textbf{0.403} & 0.364 & \textbf{0.361} & \textbf{0.358} & \textbf{0.392} & \textbf{0.330} \\
    \hline
    \hline
    \end{tabular}}

    \caption{\label{font-table} Performance on five benchmark datasets.}
    \label{results}
    \vspace{-2ex}
\end{table*}

\subsection{Dataset}
    \newcite{meng2017deep} collect a large amount of high-quality scientific metadata from various online digital libraries. We use \textbf{KP} to mark the dataset. \textbf{KP} contains 567,830 articles and we use the same splits as \newcite{meng2017deep} with 527,830 for training, 20k for validating and 20k as test dataset denoted as \textbf{KP20k}. 

    Following \newcite{meng2017deep}, we evaluate the proposed model on four widely-adopted scientific publication datasets and \textbf{KP20k} mentioned above. We take the title and abstract as the document text. In \newcite{meng2017deep}, each dataset is also split into training and test dataset for baseline models. We continue using the same test dataset for a fair comparison. Each dataset is described in detail below.

    - \textbf{Inspec} \cite{hulth2003improved}: This dataset provides
    2,000 paper abstracts. 500 testing papers adopted and their corresponding uncontrolled keyphrases for evaluation.
    
    - \textbf{Krapivin} \cite{krapivin2009large}: This dataset provides 2,304 papers with full-text and author-assigned keyphrases. We selected the first 400 papers in alphabetical order as the testing data.
    
    - \textbf{NUS} \cite{nguyen2007keyphrase}: We use
    both author-assigned and reader-assigned keyphrases and treat all 211 papers as the testing data.
    
    - \textbf{SemEval-2010} \cite{nguyen2007keyphrase}: 288 articles were collected from the ACM Digital Library. 100 articles were used for testing.
    
    - \textbf{KP20k} \cite{meng2017deep}: 567,830 scientific articles contains titles, abstracts, and keyphrases in computer science. 20k were randomly selected as test dataset.

    We use the certain part-of-speech pattern to extract the candidate phrases inspired by \newcite{le2016unsupervised}. Part-of-speech is performed by Stanford Log-linear Part-Of-Speech Tagger tools \cite{toutanova2003feature}. The following part-of-speech pattern~\cite{le2016unsupervised} is used.
\scriptsize
\begin{displaymath} 
    (JJ|JJR|JJS|VBG|VBN)*(NN|NNS|NNP|NNPS|VBG)+
\end{displaymath}
\normalsize
    The intersection of candidate phrases and keyphrases are positives denoted as $\textbf{c}^{i+}$. The number of $\textbf{c}^{i+}$ and keyphrases is shown in Table \ref{number_of_cplus_and_keyphrases}. When determining the match of two phrases, we use Porter Stemmer for preprocessing. The coverage shows that 9.42\% to 22.84\% keyphrases are not cover by candidate phrases. This will limit the final performance of our model. Our candidates only cover 77.16\% keyphrases of \textbf{NUS}, the main reason is \textbf{NUS} contains irregular keyphrases assigned by reader. The ratio of $\textbf{c}^{i-}$ and $\textbf{c}^{i+}$ is range from 9.85 to 28.85 meaning that data is imbalanced for classification task. Statistical indicators are close among \textbf{KP} dataset which shows this is a reasonable split.

\subsection{Baseline Model}
    We first compare our work with 6 baseline algorithms. Unsupervised algorithms (Tf-Idf, TextRank \cite{mihalcea2004textrank}, two-step supervised algorithms (KEA \cite{witten2005kea} and Maui \cite{medelyan2009human}) and keyphrase generation algorithms (CopyRNN \cite{chen2018keyphrase} and CorrRNN \cite{chen2018keyphrase}) are adopted. For the first 5 algorithms, we use the performance reported by \newcite{meng2017deep} and for the last one, results of \newcite{chen2018keyphrase} are used.

    We also compare with the sentence pair classification proposed by \newcite{devlin2018bert} which is shown in Figure \ref{bert_pair}. The basic setting is the same in Section \ref{bert_section} except the inputs. For sentence pair classification, the input consists of a document and a phrase with different segment embeddings. We use the final hidden vector $\textbf{x}^{D}_{[CLS]}$ corresponding to the first input token ([CLS]) as the aggregate representation. A classification layer $W \in \mathbb{R}^{d \times 2}$ is used to determine whether this phrase is a keyphrase in this document. Compared to our model, this method treats each phrase as an instance, which fails to capture the interaction between phrases and takes longer to train. Details of training time are introduced in Section \ref{training_time}.

\begin{table}[t]
    \centering
    \normalsize
    \scalebox{0.9}{
    \begin{tabular}{c|c|c c}
    \hline
    \hline
        \multirow{2}{*}{\textbf{Method}} & \textbf{Training Time} & \multicolumn{2}{|c}{\textbf{Nus}} \\
        & \textbf{hours/epoch} & $\textbf{F}_1\textbf{@5}$ & $\textbf{F}_1 \textbf{@10}$ \\
    \hline
        \textbf{BERT-Base-Pair} & 132 & 0.373 & 0.313 \\
    \hline
        \textbf{SKE-Base-Rank} & 2 & 0.381 & 0.324 \\
    \hline
        \textbf{SKE-Base-Cls} & 2 & 0.386 & 0.326 \\
    \hline
    \hline
        \textbf{BERT-Large-Pair} & - & - & -\\
    \hline
        \textbf{SKE-Large-Rank} & 7 & 0.392 & 0.328\\
    \hline
        \textbf{SKE-Large-Cls} & 7 & \textbf{0.392} & \textbf{0.330}\\
    \hline
    \hline
    \end{tabular}}

    \caption{\label{font-table} The training time of sentence pair classification and our method. As sentence pair classification model takes too much time to train based on BERT-Large, limiting its usage in practical applications, so we don't report the training time and performance.}
    \label{training_time_table}
    \vspace{-2ex}
\end{table}
\begin{table*}[t]
    \centering
    \normalsize
    \scalebox{0.775}{
    \begin{tabular}{|l|}
    \hline
    \hline
    \textbf{Title}: Deployment issues of a voip conferencing system in a virtual conferencing environment. \\
    \multirow{7}{20cm}{\textbf{Abstract}: Real time services have been supported by and large on circuitswitched networks. Recent trends favour services ported on packet switched network. For audio conferencing, we need to consider many issues scalability, quality of the conference application, floor control and load on the clients servers to name a few. In this paper, we describe an audio service framework designed to provide a virtual conferencing environment (vce). The system is designed to accommodate a large number of end users speaking at the same time and spread across the internet.  The framework is based on conference servers $\left \langle \emph{DIGIT} \right \rangle$, which facilitate the audio handling, while we exploit the sip capabilities for signaling purposes. Client selection is based on a recent quantifier called loudness number that helps mimic a physical face to face conference. We deal with deployment issues of the proposed solution both in terms of scalability and interactivity, while explaining the techniques we use to reduce the traffic. We have implemented a conference server (cs) application on a campus wide network at our institute.}
    \\ \\ \\ \\ \\ \\ \\ \\ \\
    \hline        
        \multirow{2}{20cm}{\textbf{CorrRNN}: \textbf{voip}; virtual conferencing; voip conferencing; audio conferencing; audio service; real time services; real time; distributed systems; \textbf{conference server}; \textbf{virtual conferencing environment};} \\ \\
    \hline
        \multirow{2}{20cm}{\textbf{SKE-Large-Cls}: \textbf{conference server}; \textbf{voip}; virtual conferencing; \textbf{sip}; audio conferencing;deployment; \textbf{loudness number}; scalability; \textbf{virtual conferencing environment;} conferencing;} \\ \\
    \hline
    \hline
    \end{tabular}
    }
    \caption{\label{font-table} Top10 phrases provided by CorrRNN and SKE-Large-Cls. Phrases in bold are correct. SKE-Large-Cls provides more keyphrases and better ranking results.}
    \label{case_study_table}
\end{table*}
\subsection{Training Details}
    \newcite{devlin2018bert} provides two pre-trained models named BERT-Base and BERT-Large with the same transformer structure. The main difference between them is the hidden dimension of the transformer which leads to different parameter size. BERT-Base has 110M parameters while BERT-Large has 340M parameters. We train our model both based on uncased BERT-Base and uncased BERT-Large. For both models, we use AdamW \newcite{loshchilov2017fixing} with warmup as the optimizer following. Apart from parameters of BERT, the other parameters are randomly initialized and all parameters are fine-tuned using training dataset. For the classification task, different weights for cross entropy are used for positive and negative phrases according to the ratio of them. Specifically, the weight of negative is 1 and the weight of positive is picked in \{10, 15, 20\}. For the ranking task, the margin is selected using a grid search in [0, 1]. We use the learning rate of 5e-5 and a warmup proportion of 0.1 for AdamW. The L2 weight decay rate is set to 0.01. We use minibatches of size 128 and epochs of size 5. All models are trained on a single machine with 8 x NVIDIA Tesla P40 GPUs. And the best model is selected using the validation dataset of \textbf{KP} among epochs. The random seed is fixed for a stable result.
    
    For training dataset, we only keep the documents with at least one keyphrase is matched using certain part-of-speech patterns \cite{le2016unsupervised}. 499,087 documents are left for training our proposed model.
    
    There are 42,375,099 $\left \langle \rm{document, phrase} \right \rangle$  pairs for training sentence pair classification model as this method can only process a phrase at a time which makes the training time dozens of times longer than our model. Details of training time are shown at Section \ref{training_time}. We only conduct experiments based on BERT-Base for sentence pair classification since it is not practical to use BERT-Large in this setting.

\begin{table}[t]
    \centering
    \normalsize
    \scalebox{1}{
    \begin{tabular}{c|c c}
    \hline
    \hline
        \multirow{2}{*}{\textbf{Method}}  & \multicolumn{2}{|c}{\textbf{KP20K}} \\
         & $\textbf{F}_1\textbf{@5}$ & $\textbf{F}_1 \textbf{@10}$ \\
    \hline
        \textbf{Maui} & 0.273 & 0.240 \\
    \hline
        \textbf{CopyRNN} & 0.333 & 0.262 \\
    \hline
        \textbf{SKE-RNN-Cls} & \textbf{0.339} & \textbf{0.296} \\
    \hline
    \hline
    \end{tabular}}

    \caption{\label{font-table} We conduct experiment using RNN word representation based \modelname classification model named SKE-RNN-Cls. SKE-RNN-Cls achieves 1.8\% and 12.9\% improvements over CopyRNN on F1@5 and F1@10 respectively.}
    \label{rnn_base_table}
    \vspace{-2ex}
\end{table}

\section{Results and Analysis}\label{result_section} 
    For a fair comparison, the micro-averaged F-measure is adopted following \newcite{chen2018keyphrase}.  As the standard definition, precision is defined as the number of correctly-predicted keyphrases over the number of all predicted keyphrases, and recall is computed by the number of correctly predicted keyphrases over the total number of data records. F-measure is the harmonic mean of precision and recall. 

    Table \ref{results} provides the performance of seven baseline models, as well as our proposed models. For each method, the table lists the F-measure at top 5 and top 10 results. The best scores are highlighted in bold. 

    The results show that the two unsupervised models (Tf-Idf, TextRank) have a robust performance across all datasets. The performance of the two supervised models (i.e., Maui and KEA) were unstable on some datasets. Two generation models outperform the previous baseline by a large margin, indicating the effectiveness of RNN with a copy mechanism. As no results are reported on \textbf{Inspec} and \textbf{KP20K} in \newcite{chen2018keyphrase}, we thus ignore their performance in the table. By using coverage mechanism and review mechanism, CorrRNN beats CopyRNN on three datasets. 

    We denote the sentence pair classification method based on BERT-Base as BERT-Base-Pair. BERT-Base-Pair outperforms generation method in three datasets. For \textbf{KP20k}, BERT-Base-Pair achieves 12.01\% and 19.47\% performance gain on $\textbf{F}_1\textbf{@5}$ and $\textbf{F}_1 \textbf{@10}$ respectively compared to CorrRNN. 

    We name our model with the BERT model and the training task. For example, SKE-Base-Cls and SKE-Base-Rank corresponding to the classification and ranking methods based on BERT-Base respectively. Results show our methods achieve state-of-the-art performance on all datasets except \textbf{Krapivin} datset. Our proposed method achieve at least 10\% performance gain on \textbf{NUS}, \textbf{SemEval} and \textbf{KP20k} datasets compared to generation based method. Our model also beats the sentence pair classification model on all datasets, indicating the effectiveness of the span-based feature representations.

\begin{table*}[t]
    \centering
    \normalsize
    \scalebox{0.775}{
    \begin{tabular}{|l|}
    \hline
    \hline
    \textbf{Title}: Motion estimation using modified dynamic programming \\
    \multirow{4}{20cm}{\textbf{Abstract}:
    Correspondence vector-field computation is formulated as a matching optimization problem for multiple dynamic images. The proposed method is a heuristic modification of dynamic programming applied to the 2-D optimization problem. \emph{\textbf{Motion vector field estimates}} using real movie images demonstrate good performance of the algorithm in terms of dynamic motion analysis.}
    \\ \\ \\ \\
    \hline
        \multirow{2}{20cm}{\textbf{SKE-Large-Cls}: \textbf{motion estimation}; \textbf{dynamic programming}; correspondence vector field; \emph{\textbf{motion vector field}}; \textbf{dynamic motion analysis}; \emph{\textbf{motion vector field estimates}}; \textbf{moving objects}; matching optimization; \textbf{modified dynamic programming}; motion analysis; } \\ \\ \\
    \hline
    \hline
    \textbf{Title}: A Weighted Ranking Algorithm For Facet-Based Component Retrieval System \\
    \multirow{6}{20cm}{\textbf{Abstract}:
    Facet-based component retrieval techniques have been proved to be an effective way for retrieving. These Techniques are widely adopted by component library systems, but they usually simply list out all the retrieval results without any kind of ranking. In our work, we focus on the problem that how to determine the ranks of the components retrieved by user. Factors which can influence the ranking are extracted and identified through the analysis of ER-Diagram of facet-based component library system. In this paper, a mathematical model of \emph{\textbf{weighted ranking algorithm}} is proposed and the timing of ranks calculation is discussed. Experiment results show that this algorithm greatly improves the efficiency of component retrieval system..}
    \\ \\ \\ \\ \\ \\
    \hline
        \multirow{2}{20cm}{\textbf{SKE-Large-Cls}: \textbf{component retrieval}; \textbf{facet}; weighted ranking; ranks; er diagram; \emph{\textbf{weighted ranking algorithm}}; \textbf{component library}; component library system; \emph{\textbf{ranking algorithm}}; \textbf{component retrieval system};} \\ \\ 
    \hline
    \hline
    \end{tabular}
    }
    \caption{\label{font-table} Top10 phrases provided by SKE-Large-Cls. Phrases in bold are correct. Phrases in italic and bold are overlapped keyphrases and corresponding texts are also bold and italic in the document above.}
    \label{case_study_table_2}
    \vspace{-2ex}
\end{table*}
\subsection{Training Time}\label{training_time}
    We show the training time of sentence pair classification and our method in Table \ref{training_time_table}. Our models achieve significant performance gain and are 66 times faster than sentence pair classification method which is proportional to the size of the training dataset. As the sentence pair classification model takes too much time to train based on BERT-Large, limiting its usage in practical applications, so we don't report the training time and performance.

\subsection{Word Representation with RNN}\label{word_rep_with_rnn}
    We also conduct a experiment using Glove~\cite{pennington2014glove} with RNN word representation other than BERT. We compared this model with CopyRNN on the largest test dataset \text{KP20K}. Result is shown in Table \ref{rnn_base_table}. SKE-RNN-Cls achieves  1.8\% and 12.9\% improvements over CopyRNN on F1@5 and F1@10 respectively. 
    
\subsection{Overlapped Keyphrases}\label{overlap_keyphrases}
	To better evaluate our model’s ability of identifying overlapped keyphrses, we make some statics on the testing dataset. There are 16.67\% and 19.96\% overlapped keyphrase in the \textbf{NUS} and \textbf{SemEval}, respectively. The results show that our SKE-Large-CLS model could extract 69.70\% and 73.77\% of the overlapped keyphrases in these two datasets, which illustrates the ability of our model on extracting overlapped keyphrases.

\section{Discussion}\label{discussion}
\subsection{Case Study}\label{case_study}
    We compare the phrases provided by CorrRNN and SKE-Large-Cls on an example article shown in Table \ref{case_study_table}. Compared to CorrRNN, SKE-Large-Cls provided two more keyphrases ``sip'' and ``loudness number'' which cover two important topics. Moreover, SKE-Large-Cls presents a better ranking of keyphrases(i.e. SKE-Large-Cls provides three keyphrases at Top4 results while CorrRNN only provides one). Though using coverage mechanism and review mechanism, CorrRNN still produces continuous phrases that have the same prefix(i.e. ``audio conferencing'' and  ``audio service'', ``real time services'' and ``real time''), showing that generation method fails to obtain the semantic meanings of phrases as a whole.

    We show another example in Table \ref{case_study_table_2} in which SKE-Large-Cls provided two overlapped keyphrases. For the first example, ``motion vector field estimates'' and ``motion vector field'' both are keyphrases and overlapped in the document. The same situation appeared in the second example with keyphrases ``weighted ranking algorithm'' and ``ranking algorithm''. SKE-Large-Cls obtains them all, showing the ability to process the overlapped keyphrases while this situation is hard to handle using the sequence labeling method.
\subsection{Applicable Task}\label{applicable_tasks_section}
    In this paper, we mainly focus on keyphrase extraction task, whereas our proposed method is applicable to other natural language tasks such as named entity recognition. For named entity recognition task, existing method\cite{ling2012fine} uses a sequence labeling model to tagging the candidate entity using ``B'', ``I'', ``O'' and another multi-class multi-label classification model to type the entity, e.g. organization, person. However, sequence labeling can hardly solve the overlapped name entities (e.g. ``BMW X1'' and ``BMW''  are a car brand and a car series respectively, showing different granularity of information) which can be solved using our method. Furthermore, by changing the candidate selection stage in this paper, we can obtain an end-to-end method for keyphrase extraction and named entity recognition. For example, given a document, we can split it by symbols that hardly appear in an entity, such as comma, period. Then, all possible spans in the splits are selected as candidates. 

\section{Conclusion and Future Work}
\label{ConclusionAndFutureWork}

The key idea of \modelname model is to build span-based feature representation for keyphrases. Span-based approach is capable of tackling overlapping keyphrases effectively. Furthermore, our proposed model is able to utilize context information based on an overall understanding of the document and a better considering of interactions between phrases by Bi-LSTM models. 
Comprehensive empirical studies demonstrate the effectiveness of our proposed model on four benchmark datasets. 


Future works focus on three aspects: (i) Our model adopt two-step approach, so it cannot avoid error accumulation. Motivated by this, We are pursuing a method to improve the coverage (e.g. a more exhaustive part-of-speech pattern, using all span limited by a threshold length as candidate phrases and using all possible spans in segment split by symbols). 
(ii) We adopt our model in other natural language processing tasks, (e.g. named entity recognition or automatic text summarization) which need to extract continuous texts for classification or ranking. 
(iii) Our proposed model views the same phrases appear in different positions as difference phrases and fails to utilize this valuable information. It would be interesting to explore a new model structure to solve this problem.

    
    

\clearpage
\bibliographystyle{aaai}
\bibliography{span_bib}
\end{document}